\documentclass[letterpaper]{article} 
\usepackage{aaai24}  
\usepackage{times}  
\usepackage{helvet}  
\usepackage{courier}  
\usepackage[hyphens]{url}  
\usepackage{graphicx} 
\usepackage{natbib}  
\usepackage{caption} 
\usepackage{algorithm}
\usepackage{algpseudocode}

\usepackage{amsmath, amsthm, amssymb}
\usepackage{booktabs} 
\usepackage{multirow}
\usepackage{caption}
\usepackage{lipsum}
\usepackage{subcaption}

\theoremstyle{plain}
\newtheorem{theorem}{Theorem}

\theoremstyle{definition}

\usepackage{xcolor}
\newcommand{\fref}[1]{Fig.~\ref{#1}}
\newcommand{\tref}[1]{Table~\ref{#1}}

\usepackage{newfloat}
\usepackage{listings}
\DeclareCaptionStyle{ruled}{labelfont=normalfont,labelsep=colon,strut=off} 
\floatstyle{ruled}
\newfloat{listing}{tb}{lst}{}
\floatname{listing}{Listing}
\title{CR-SAM: Curvature Regularized Sharpness-Aware Minimization}
\author{
    Tao Wu\textsuperscript{\rm 1},
    Tie Luo\textsuperscript{\rm 1}\thanks{Corresponding author.}, 
    Donald C. Wunsch II\textsuperscript{\rm 2}
}
\affiliations{
    \textsuperscript{\rm 1}Department of Computer Science, Missouri University of Science and Technology\\
    \textsuperscript{\rm 2}Department of Electrical and Computer Engineering, Missouri University of Science and Technology\\

    \{wuta, tluo, dwunsch\}@mst.edu
}

\usepackage{bibentry}

\begin{document}

\maketitle

\begin{abstract}
The capacity to generalize to future unseen data stands as one of the utmost crucial attributes of deep neural networks. Sharpness-Aware Minimization (SAM) aims to enhance the generalizability by minimizing worst-case loss using one-step gradient ascent as an approximation. However, as training progresses, the non-linearity of the loss landscape increases, rendering one-step gradient ascent less effective. On the other hand, multi-step gradient ascent will incur higher training cost. In this paper, we introduce a normalized Hessian trace to accurately measure the curvature of loss landscape on {\em both} training and test sets. In particular, to counter excessive non-linearity of loss landscape, we propose Curvature Regularized SAM (CR-SAM), integrating the normalized Hessian trace as a SAM regularizer. Additionally, we present an efficient way to compute the trace via finite differences with parallelism. Our theoretical analysis based on PAC-Bayes bounds establishes the regularizer's efficacy in reducing generalization error. Empirical evaluation on CIFAR and ImageNet datasets shows that CR-SAM consistently enhances classification performance for ResNet and Vision Transformer (ViT) models across various datasets. Our code is available at https://github.com/TrustAIoT/CR-SAM.
\end{abstract}

\section{Introduction}
\label{sec:Introduction}

Over the past decade, rapid advancements in deep neural networks (DNNs) have significantly reshaped various pattern recognition domains including computer vision \cite{he2016deep}, speech recognition \cite{oord2018parallel}, and natural language processing \cite{kenton2019bert}. However, the success of DNNs hinges on their capacity to generalize---how well they would perform on new, unseen data. With their intricate multilayer structures and non-linear characteristics, modern DNNs possess highly non-convex loss landscapes that remain only partially understood. Prior landscape analysis has linked flat local minima to better generalization \cite{hochreiter1997flat,keskar2016large,dziugaite2017computing,neyshabur2017exploring,Jiang2020Fantastic}. In particular, \cite{Jiang2020Fantastic} conducted a comprehensive empirical study on various generalization metrics, revealing that measures based on sharpness exhibit the highest correlation with generalization performance. \cite{dziugaite2017computing} computed the generalization error to elucidate the effective generalization of over-parameterized DNNs trained through stochastic gradient descent (SGD). Recently, \cite{foret2021sharpnessaware} introduced Sharpness-Aware Minimization (SAM), along with an efficient technique for minimizing loss landscape sharpness. This method has proven highly effective in enhancing DNN generalization across diverse scenarios. Given SAM's remarkable success and the significance of DNN generalization, a substantial body of subsequent research has emerged \cite{liu2022towards,du2022efficient,mi2022make,du2022sharpness,jiang2023an,liu2022random}.

Specifically, the SAM approach formulates the optimization of neural networks as a minimax problem, where it aims to minimize the maximum loss within a small radius $\rho$ around the parameter $\boldsymbol{w}$. Given that the inner maximization problem is NP-hard, SAM employs a practical sharpness calculation method that utilizes one-step gradient ascent as an approximation.

However, our experimentation reveals a notable decline in the accuracy of this one-step approximation as training progresses (for a glimpse, refer to \fref{fig:AR}). This phenomenon likely stems from the heightened non-linearity within the loss landscape during later stages of training. Our further investigation highlights a limitation in conventional curvature measures like the Hessian trace and the top eigenvalue of the Hessian matrix. These measures diminish as training advances, incorrectly suggesting reduced curvature and overlooking the actual non-linear characteristics.

Consequently, we posit that the escalating non-linearity in SAM training undermines the precision of approximating and effectiveness of mitigating sharpness. Building upon these insights, we introduce the concept of a {\em normalized Hessian trace}. This novel metric serves as a dependable indicator of loss landscape non-linearity and behaves consistently across training and testing datasets. Guided by this metric, we propose {\em Curvature Regularized SAM} (CR-SAM), a novel regularization approach for SAM training. CR-SAM incorporates the normalized Hessian trace to counteract excessive non-linearity effectively.

To calculate the normalized Hessian trace, we present a computationally efficient strategy based on finite differences (FD). This approach enables parallel execution without additional computational burden. Through both theoretical analysis and empirical evaluation, we demonstrate that CR-SAM training converges towards flatter minima, resulting in substantially enhanced generalization performance.

Our main contributions can be summarized as follows: 
\begin{itemize}
    \item We identify that the one-step gradient ascent approximation becomes less effective during the later stages of SAM training. In response, we introduce normalized Hessian trace, a metric that can accurately and consistently characterize the non-linearity of neural network loss landscapes.

    \item We propose CR-SAM, a novel algorithm that infuses curvature minimization into SAM and thereby enhance the generalizability of deep neural networks. For scalable computation, we devise an efficient technique to approximate the Hessian trace using finite differences (FD). This technique involves only independent function evaluations and can be executed in parallel without additional overhead. Moreover, we also theoretically show the efficacy of CR-SAM in reducing generalization error, leveraging PAC-Bayes bounds.

    \item Our comprehensive evaluation of CR-SAM spans a diverse range of contemporary DNN architectures. The empirical findings affirm that CR-SAM consistently outperforms both SAM and SGD in terms of improving model generalizability, across multiple datasets including CIFAR10/100 and ImageNet-1k/-C/-R.
\end{itemize}

\section{Background and Related Work}
\label{sec:Background}

Empirical risk minimization (ERM) is a fundamental principle in machine learning for model training on observed data.  Given a training dataset $\mathcal{S} = \left\{\left(x_i, y_i\right)\right\}_{i=1}^n$ drawn i.i.d. from an underlying unknown distribution $\mathcal{D}$, we denote by $f(x ; \boldsymbol{w})$ a deep neural network model with trainable parameters $\boldsymbol{w} \in \mathbb{R}^p$, where a differentiable loss function w.r.t. an input $x_i$ is given by $\ell\left(f\left(x_i ; \boldsymbol{w}\right), y_i\right)$ and is taken to be the cross entropy loss in this paper. 
The \emph{empirical loss} can be written as $L_{\mathcal{S}}(\boldsymbol{w}) = \frac{1}{n}\sum_{i=1}^n \ell\left(f\left(x_i ; \boldsymbol{w}\right), y_i\right)$ whereas the \emph{population loss} is defined as $L_{\mathcal{D}}(\boldsymbol{w}) = \mathbb{E}_{(x, y) \sim D}[\ell\left(f\left(x ; \boldsymbol{w}\right), y\right)]$. The generalization error is defined as the difference between $L_{\mathcal{D}}(\boldsymbol{w})$ and $L_{\mathcal{S}}(\boldsymbol{w})$, i.e., $e(f)=L_{\mathcal{D}}(\boldsymbol{w})-L_{\mathcal{S}}(\boldsymbol{w})$.

\subsection{SAM and Variants}
\label{subsec:sam}
Sharpness-Aware Minimization (SAM) \cite{foret2021sharpnessaware} is a novel optimization algorithm that directs the search for model parameters within flat regions. Training DNNs with this method has demonstrated remarkable efficacy in enhancing generalization, especially on transformers. SAM introduces a new objective that aims to minimize the maximum loss in the vicinity of weight $\boldsymbol{w}$ within a radius $\rho$:
\begin{equation*}
\min _{\boldsymbol{w}} L^{\operatorname{SAM}}(\boldsymbol{w})\; \text { where }\; L^{\operatorname{SAM}}(\boldsymbol{w}) = \max _{\|\boldsymbol{v}\|_2 \leq 1} L_{\mathcal{S}}(\boldsymbol{w}+\rho \boldsymbol{v}).
\end{equation*}

Through the minimization of the SAM objective, the neural network's weights undergo updates that shift them towards a smoother loss landscape. As a result, the model's generalization performance is improved. To ensure practical feasibility, SAM adopts two approximations: (1) employs one-step gradient ascent to approximate the inner maximization; (2) simplifies gradient calculation by omitting the second and higher-order terms, i.e.,
\begin{equation}\label{eq:sam-approx}
\nabla L^{\operatorname{SAM}}(\boldsymbol{w}) \approx \nabla L_{\mathcal{S}}\left(\boldsymbol{w}+\rho \frac{\nabla L_{\mathcal{S}}(\boldsymbol{w})} {\|\nabla L_{\mathcal{S}}(\boldsymbol{w})\|_2}\right).
\end{equation}

Nevertheless, behind the empirical successes of SAM in training computer vision models \cite{foret2021sharpnessaware,chen2022when} and natural language processing models \cite{bahri2021sharpness}, there are two inherent limitations.

Firstly, SAM introduces a twofold computational overhead to the base optimizer (e.g., SGD) due to the inner maximization process. In response, recent solutions such as LookSAM \cite{liu2022towards}, Efficient SAM (ESAM) \cite{du2022efficient}, Sparse SAM (SSAM) \cite{mi2022make}, Sharpness-Aware Training for Free (SAF) \cite{du2022sharpness}, and Adaptive policy SAM (AE-SAM) \cite{jiang2023an} have emerged, which propose various strategies to reduce the added overhead. 

Secondly, the high non-linearity of DNNs' loss landscapes means that relying solely on one-step ascent may not consistently lead to an accurate approximation of the maximum loss. To address this issue, Random SAM (R-SAM) \cite{liu2022random} introduced random smoothing to the loss landscape, but its reliance on heuristic methods without strong theoretical underpinnings limits its justification. Empirically, we have included R-SAM in our baseline comparisons, demonstrating our method's superior performance (as seen in \tref{tab:imagenet}).

Beyond these two limitations, our proposed approach to enhance computational efficiency remains orthogonal to the various SAM variants mentioned above. It can be seamlessly integrated with them, further amplifying efficiency gains.

\begin{figure*}[t!]
   \centering
   \begin{subfigure}[b]{0.4\textwidth}
     \centering
     \includegraphics[width=\textwidth]{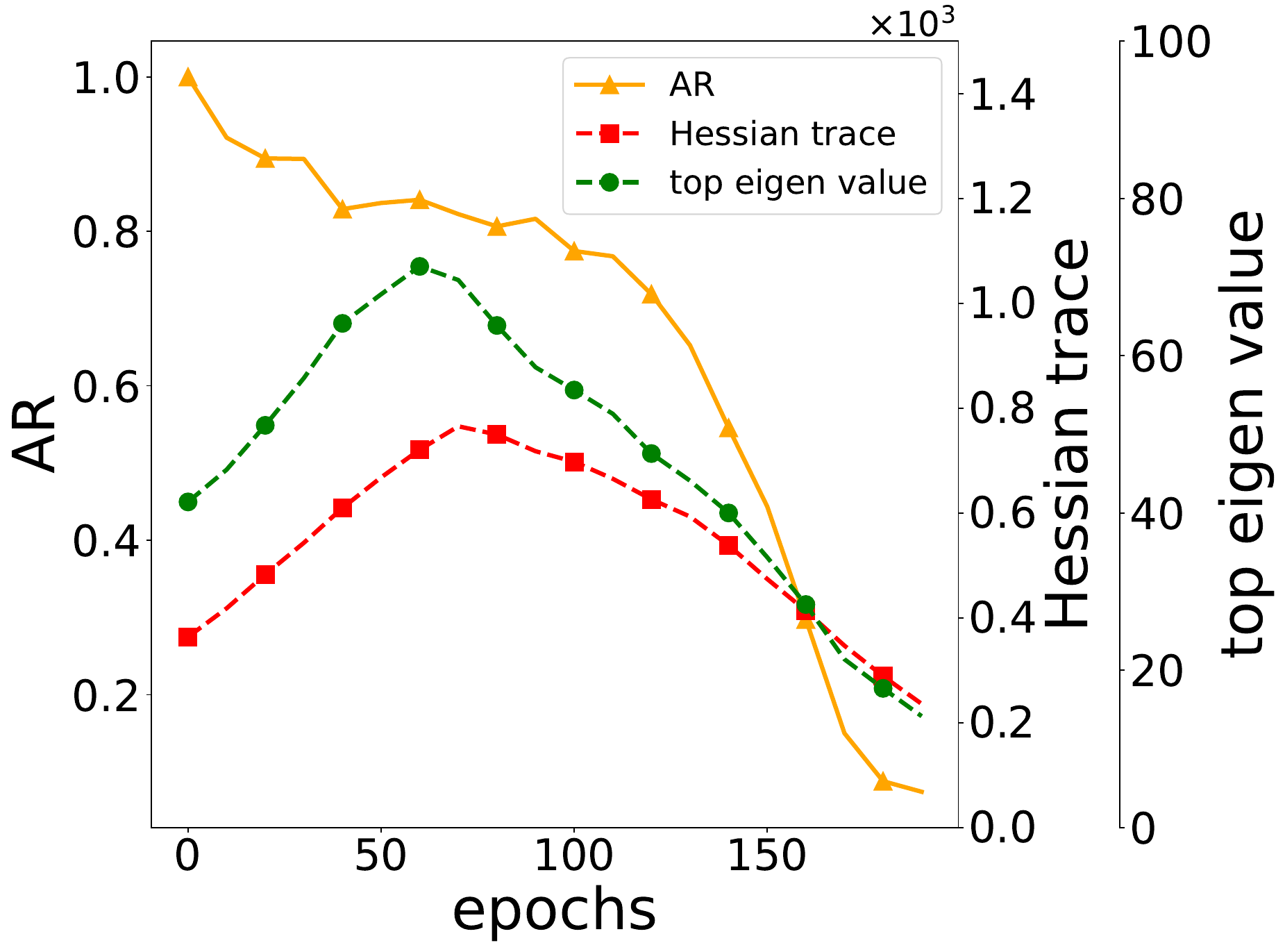}
     \caption{ResNet-18 on CIFAR10}
     \label{subfig:cifar10}
   \end{subfigure}
   \hspace{0.05\textwidth}
   \begin{subfigure}[b]{0.4\textwidth}
     \centering
     \includegraphics[width=\textwidth]{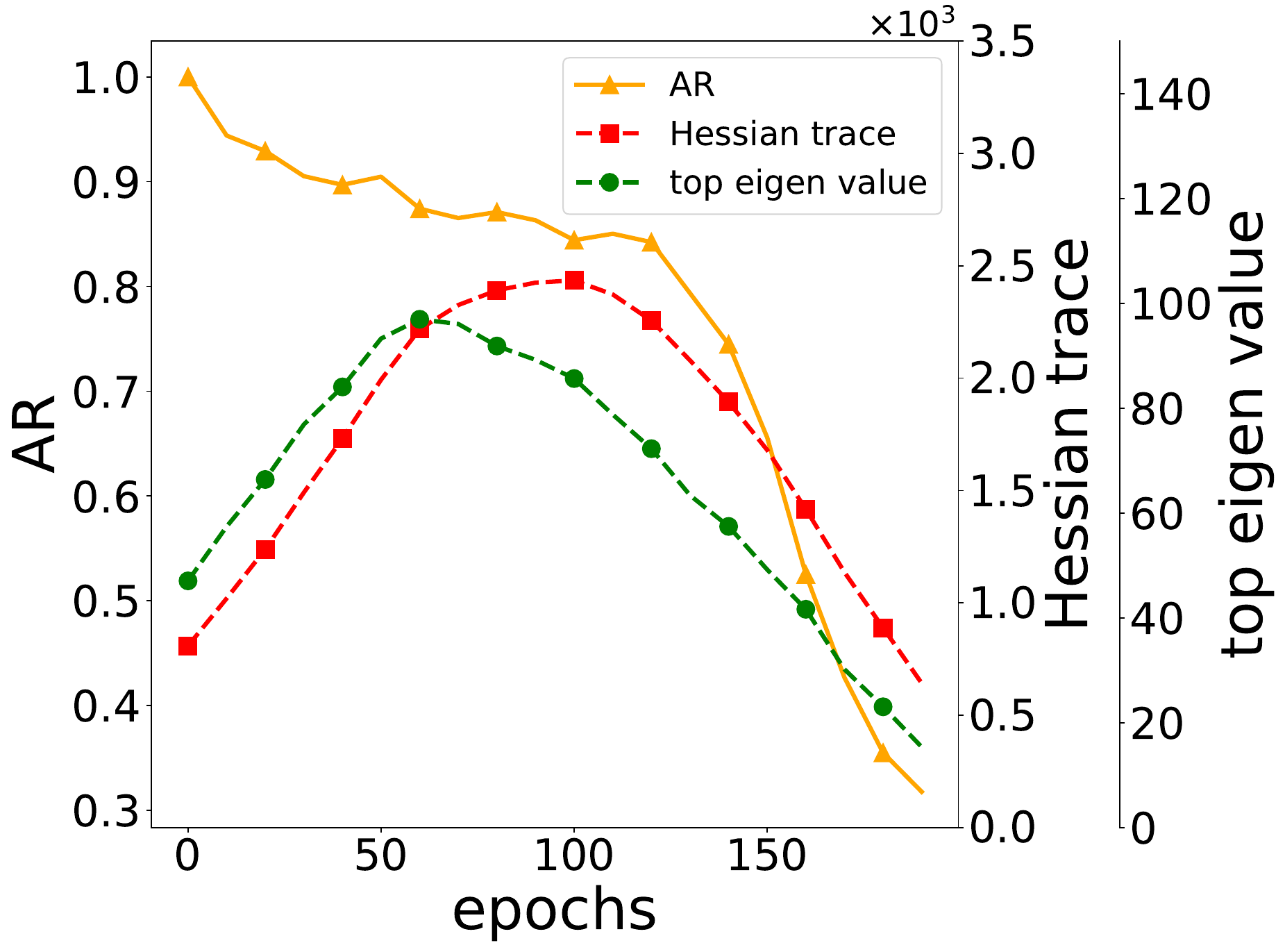}
     \caption{ResNet-18 on CIFAR100}
     \label{subfig:cifar100}
   \end{subfigure}
\caption{\footnotesize The evolution of approximation ratio (AR), Hessian trace and top eigenvalue of Hessian (the two Y axes on the right) during SAM training on CIFAR10 and CIFAR100 datasets. The continuously decreasing AR indicates an enlarging curvature whereas both of the Hessian-based curvature metrics (which are expected to continuously increase) fail to capture the true curvature of model loss landscape.}
\label{fig:AR}
\vspace{-0.5cm}
\end{figure*}

\subsection{Regularization Methods for Generalization}
\label{subsec:regularization}
The work \cite{NEURIPS2020_322f6246} contends that model generalization hinges primarily on two traits: the model's \emph{support} and its \emph{inductive biases}. Given the broad applicability of modern DNNs to various datasets, the inductive biases is the remaining crucial factor for guiding a model towards the true data distribution. From a Bayesian standpoint, inductive bias can be viewed as a prior distribution over the parameter space. Classical $\ell_1$ and $\ell_2$ regularization, for instance, correspond to Laplacian and Gaussian prior distributions respectively. In practice, one can employ regularization techniques to instill intended inductive biases, thereby enhancing model generalization. Such regularization can be applied to three core components of modern deep learning models: data, model architecture, and optimization.

{\bf Data-based regularization} involves transforming raw data or generating augmented data to combat overfitting. Methods like label smoothing \cite{szegedy2016rethinking}, Cutout \cite{devries2017improved}, Mixup \cite{zhang2017mixup}, and RandAugment \cite{cubuk2020randaugment} fall under this category. {\bf Model-based regularization} aids feature extraction and includes techniques such as dropout \cite{srivastava2014dropout}, skip connections \cite{he2016deep}, and batch normalization \cite{ioffe2015batch}. Lastly, {\bf optimization-based regularization} imparts desired properties like sparsity or complexity into the model. Common methods include weight decay \cite{krogh1991simple}, gradient norm penalty \cite{drucker1992improving, zhao2022penalizing}, Jacobian regularization \cite{sokolic2017robust}, and confidence penalty \cite{pereyra2017regularizing}. Our proposed curvature regularizer in this work aligns with the optimization-based strategies, fostering flatter loss landscapes.

\subsection{Flat Minima}
\label{subsec:flat}
Recent research into loss surface geometry underscores the strong correlation between generalization and the flatness of minima reached by DNN parameters. Among various mathematical definitions of flatness, including $\epsilon$-sharpness \cite{keskar2016large}, PAC-Bayes measure \cite{Jiang2020Fantastic}, Fisher Rao Norm \cite{liang2019fisher}, and entropy measures \cite{pereyra2017regularizing,chaudhari2019entropy}, notable ones include Hessian-based metrics like Frobenius norm \cite{wu2022alignment, wu2022does}, trace of the Hessian \cite{dinh2017sharp}, largest eigenvalue of the Hessian \cite{kaur2023maximum}, and effective dimensionality of the Hessian \cite{maddox2020rethinking}. In this work, our focus is on exploring the Hessian trace and its connection to generalization. Akin to our objective, \cite{LIU202313} also proposes Hessian trace regularization for DNNs. However, \cite{LIU202313} utilizes the computationally demanding Hutchinson method \cite{avron2011randomized} with dropout as an unbiased estimator for Hessian trace. In contrast, our method employs finite difference (FD), offering greater computational efficiency and numerical stability. Moreover, our rationale and regularization approach significantly differ from \cite{LIU202313}.

\begin{figure*}[ht!]
   \centering
   \begin{subfigure}{0.45\textwidth}
     \centering
     \includegraphics[width=\textwidth]{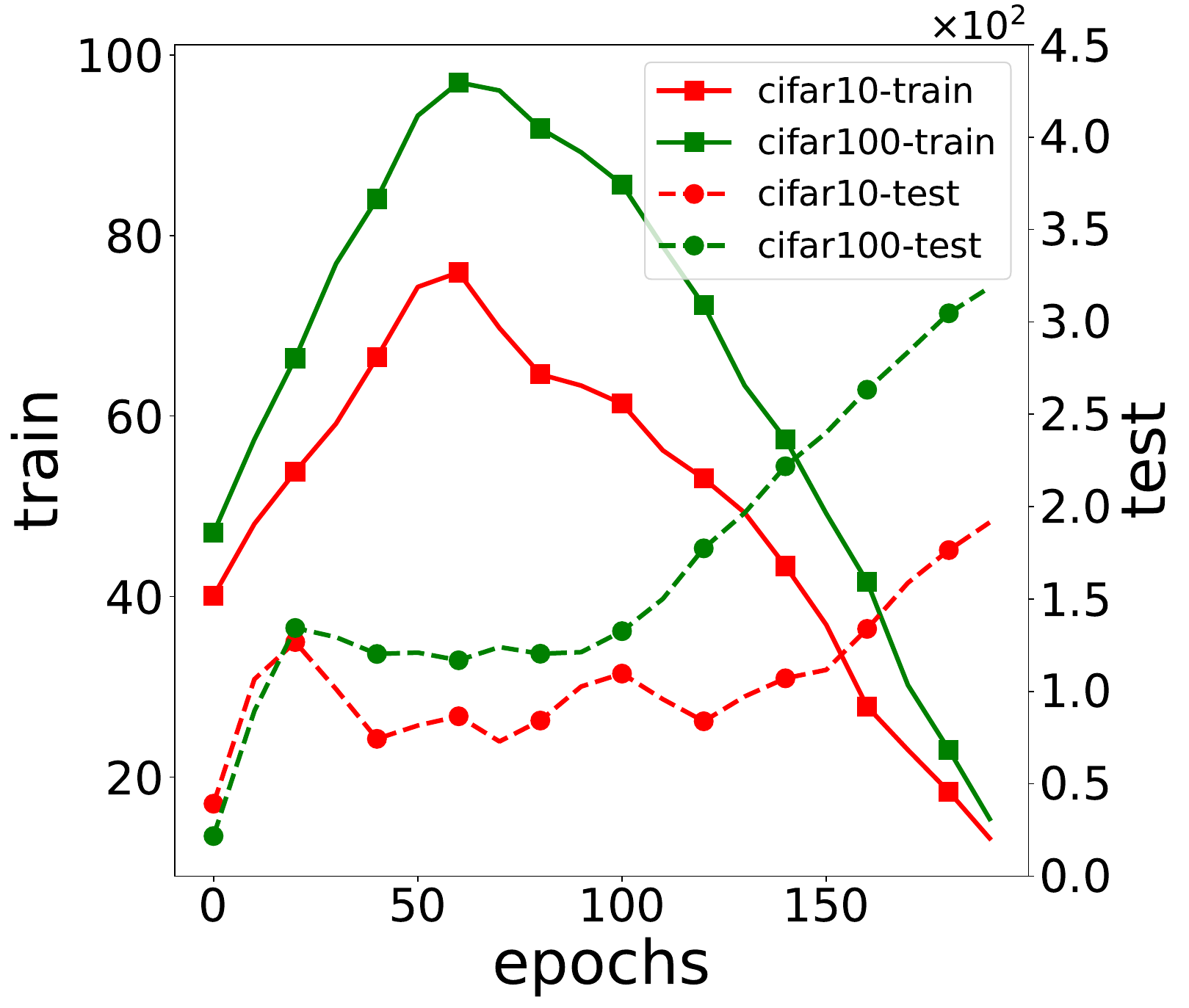}
     \caption{$\lambda_{top}\left(\nabla^2 L_{\mathcal{S}}\left(\boldsymbol{w}\right)\right)$}
     \label{subfig:top_eigen}
   \end{subfigure}
   \hspace{0.01\textwidth}
   \begin{subfigure}{0.45\textwidth}
     \centering
     \includegraphics[width=\textwidth]{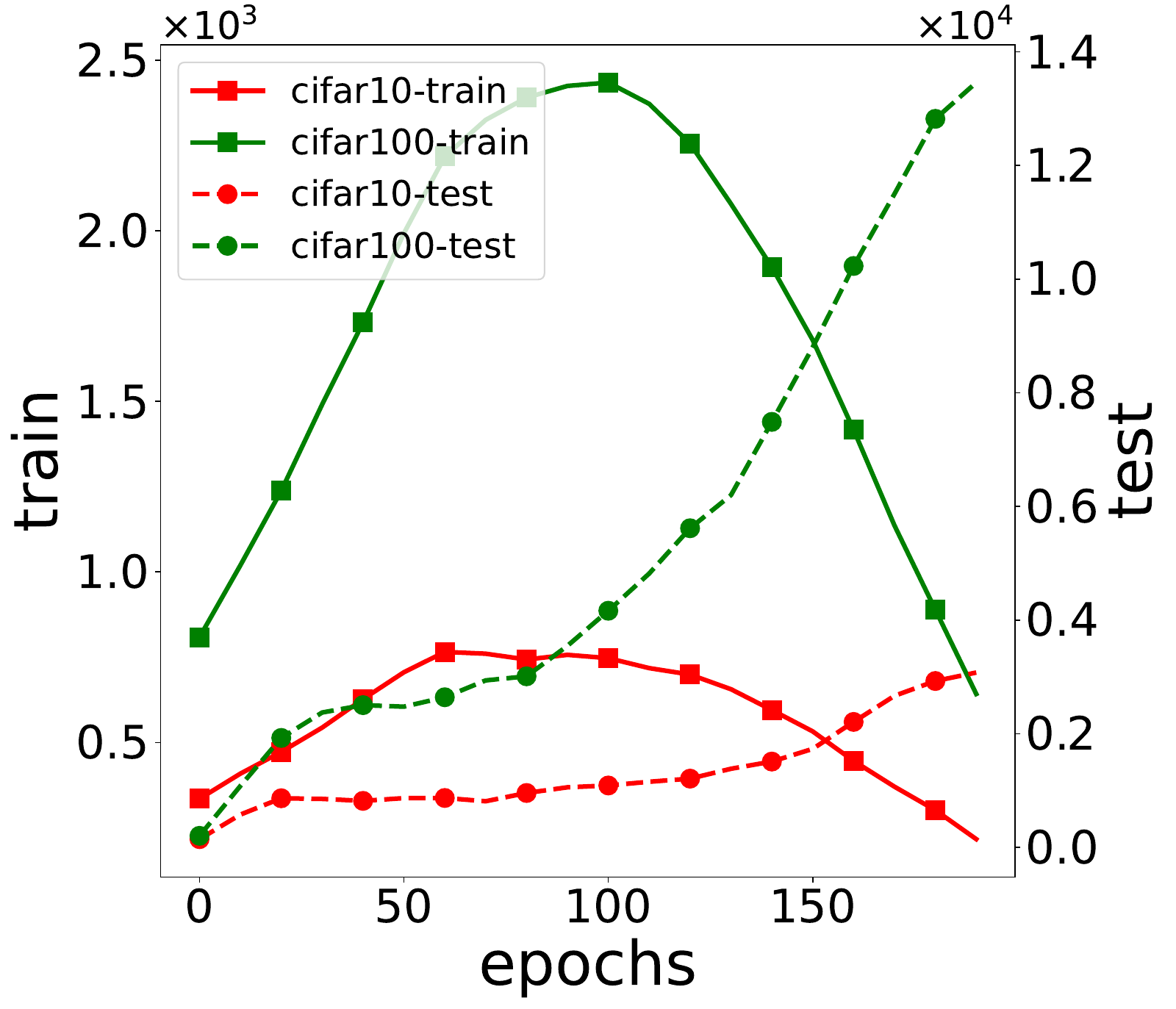}
     \caption{$\operatorname{Tr}\left(\nabla^2 L_{\mathcal{S}}\left(\boldsymbol{w}\right)\right)$}
     \label{subfig:trace}
   \end{subfigure}
   \hfill
   \begin{subfigure}{0.45\textwidth}
     \centering
     \includegraphics[width=\textwidth]{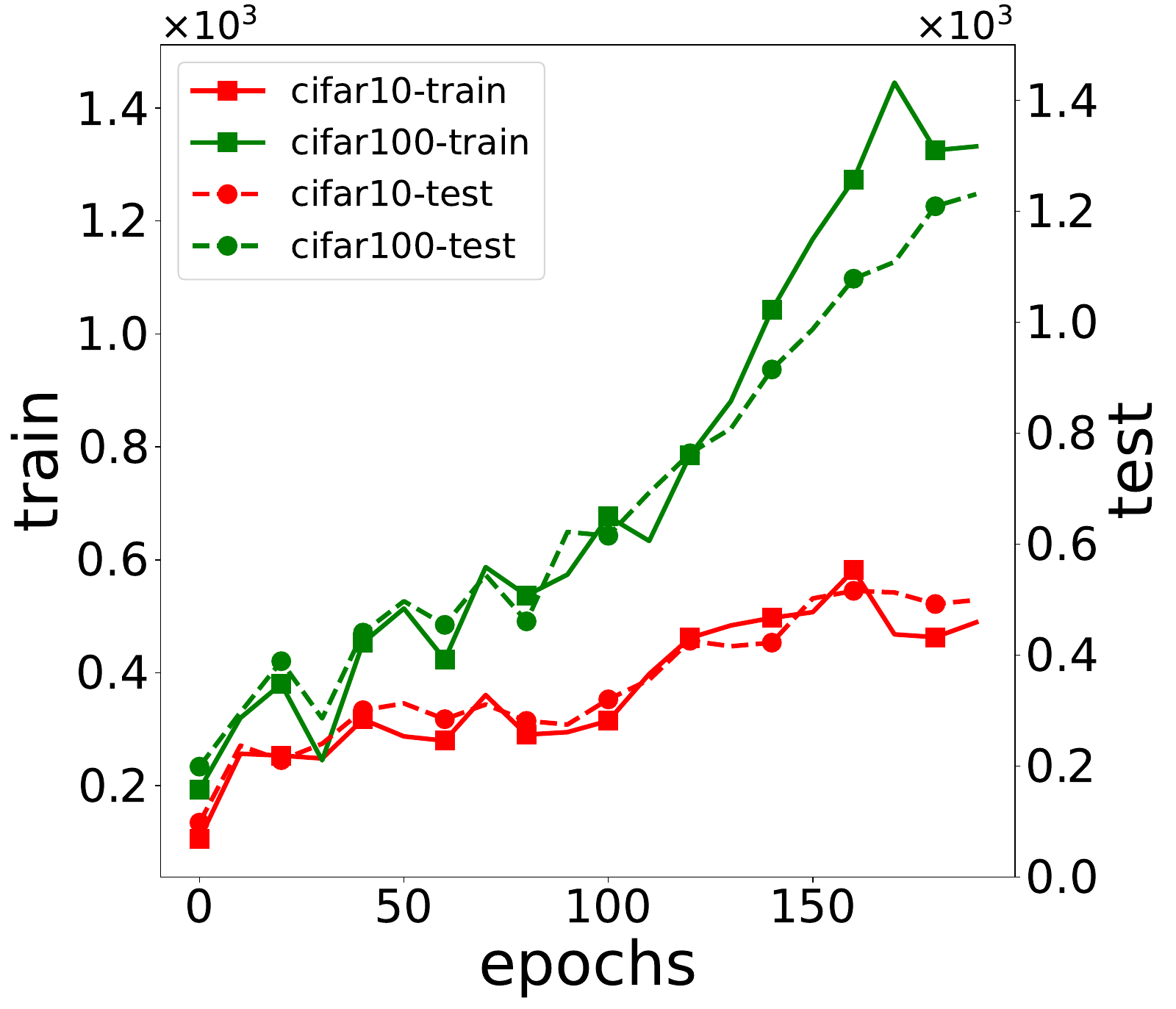}
     \caption{$\mathcal{C}(\boldsymbol{w})$}
     \label{subfig:n_trace}
   \end{subfigure}
    \hspace{0.01\textwidth}
   \begin{subfigure}{0.45\textwidth}
     \centering
     \includegraphics[width=\textwidth]{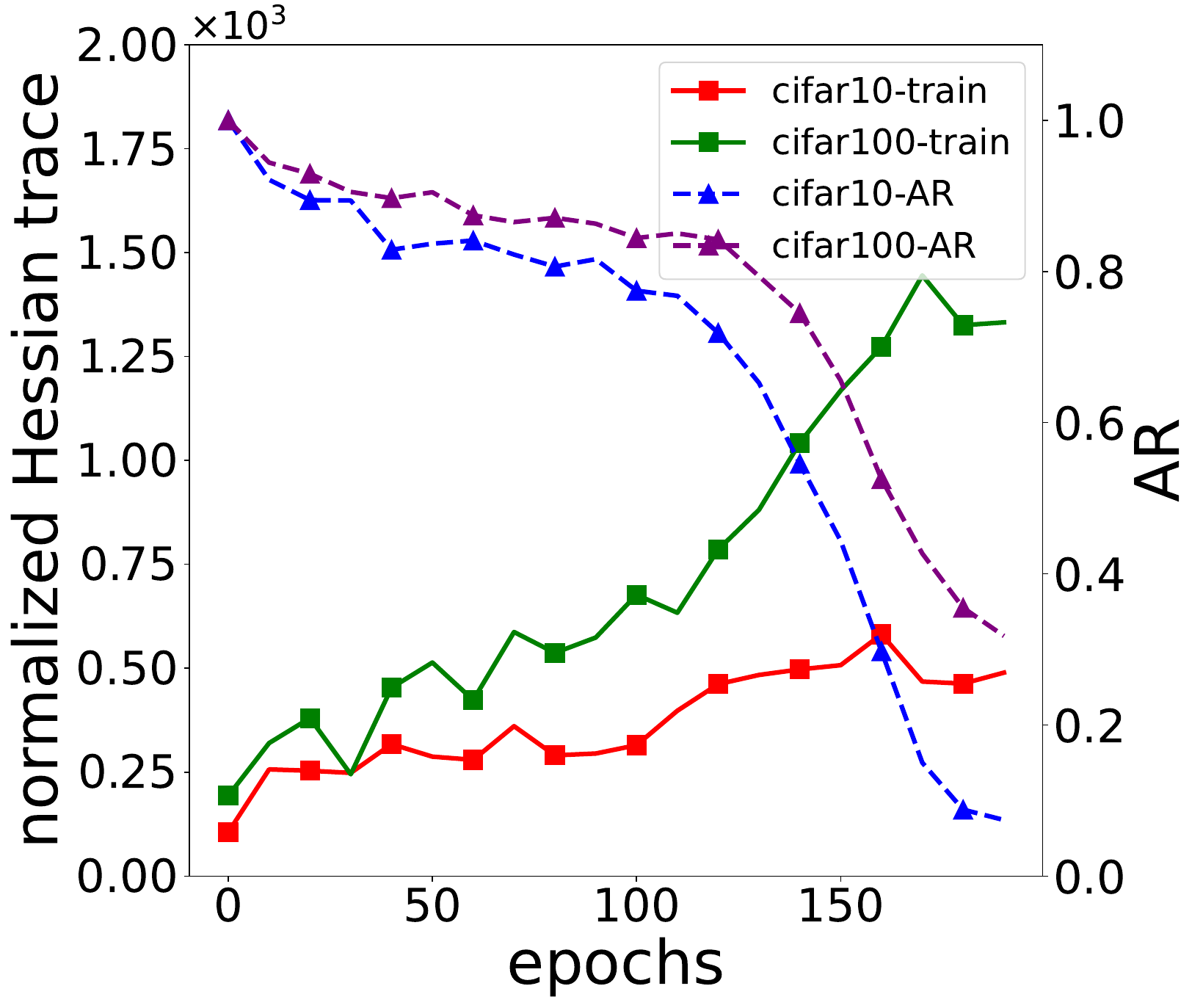}
     \caption{$\mathcal{C}(\boldsymbol{w})  \: v.s. \: \operatorname{AR}$}
     \label{subfig:n_trace_vs_ar}
   \end{subfigure}
    \caption{Evolution of the three curvature metrics (indicated in the captions of subfig.(a)(b)(c)) during SAM training of ResNet-18 on CIFAR-10 and CIFAR-100. In (a)(b)(c), the left/right Y axes denote the metric values on training/test sets, also corresponding to solid/dashed lines. Subfigs. (a) (b) show that the top Hessian eigenvalue and Hessian trace exhibit large discrepancy on train and test sets where values calculated on test set can be 50x more than those on training set. Subfig (c) shows that our proposed normalized Hessian trace shows consistent trends which implies that it well captures the true model geometry. Finally, subfig (d) illustrates that the normalized Hessian trace also reflects (inversely) the phenomenon of decreasing approximation ratio (AR) since they both indicate a growing curvature throughout training.}
    \label{fig:CR}
\end{figure*}

\section{Methodology}
\label{sec:Methodology}

\subsection{Our Empirical Findings about SAM Training}\label{subsec:empirical}

\textbf{1) Declining accuracy of one-step approximation.} 
The optimal solution to the inner maximization in SAM's objective is intractable, which led SAM to resort to an approximation using one-step gradient ascent. However, we found that this approximation's accuracy diminishes progressively as training advances. To show this, we introduce the \emph{approximation ratio (AR)} for sharpness, approximated by one-step gradient ascent, defined as:
\vspace{-0.5cm}
\begin{equation}
\operatorname{AR} = \mathbb{E}_{(x, y) \sim D}\left[\frac{\ell\left(f\left(x; \boldsymbol{w+\delta}\right), y\right)-\ell\left(f\left(x ; \boldsymbol{w}\right), y\right)}{\ell\left(f\left(x; \boldsymbol{w+\delta^*}\right), y\right)-\ell\left(f\left(x ; \boldsymbol{w}\right), y\right)}\right]
\end{equation}
where $\boldsymbol{\delta}$ represents one-step gradient ascent perturbation, and $\boldsymbol{\delta^*}$ denotes the optimal perturbation. An AR closer to 1 indicates a better approximation. Given the infeasibility of obtaining the optimal $\boldsymbol{\delta^*}$, we employ the perturbation from a 20-step gradient ascent as $\boldsymbol{\delta^*}$ and approximate its expectation by sampling 5000 data points from the training set and calculating their average. Our assessment of AR through multiple experiments, illustrated in \fref{fig:AR}, reveals its progression during training. Notably, the one-step ascent approximation for sharpness demonstrates diminishing accuracy as training unfolds, with a significant decline in the later stages. This suggests an increasing curvature of the loss landscape as training advances. In the realm of DNNs, the curvature of a function at a specific point is commonly assessed through the Hessian matrix calculated at that point. However, the dependence on gradient scale make Hessian metrics fail to measure the curvature precisely. Specifically, models near convergence of training exhibit smaller gradient norms and inherently correspond to reduced Hessian norms, but does not imply a more linear model.

We show the evolution of conventional curvature metrics like Hessian trace and the top eigenvalue of the Hessian in \fref{fig:AR}, both metrics increase initially and then decrease, which fail to capture true loss landscape curvature since AR's consistent decline implies a higher curvature. This phenomenon also verify their dependence on the scaling of model gradients; as gradients decrease near convergence, Hessian-based curvature metrics like Hessian trace and top eigenvalue of the Hessian also decrease.

The degrading effectiveness of the one-step gradient ascent approximation can be theoretically confirmed through a Taylor expansion. The sharpness optimized by SAM in practice is represented as:
\begin{align}\label{eq:taylor}
R^{\mathrm{SAM}}(\boldsymbol{w})&=L_{\mathcal{S}}\left(\boldsymbol{w}+\rho \frac{\nabla L_{\mathcal{S}}(\boldsymbol{w})}{\|\nabla L_{\mathcal{S}}(\boldsymbol{w})\|_2}\right)-L_{\mathcal{S}}(\boldsymbol{w}) \notag \\
&=\rho\|\nabla L_{\mathcal{S}}(\boldsymbol{w})\|_2 + O\left(\rho^2\right)
\end{align}
Eq.~\eqref{eq:taylor} highlights that as training nears convergence, the gradient $\nabla L_{\mathcal{S}}(\boldsymbol{w})$ tends toward $0$, causing $R^{\mathrm{SAM}}(\boldsymbol{w})$ to approach $0$ as well. Consequently, sharpness ceases to be effectively captured, and SAM training mirrors standard training behavior.

\textbf{2) A new metric for accurate curvature characterization.} 
Our initial observation underscores the limitations of the top Hessian eigenvalue and Hessian trace in capturing loss landscape curvature during SAM training. These metrics suffer from sensitivity to gradient scaling, prompting the need for a more precise curvature characterization. To address this challenge, we introduce a novel curvature metric, \emph{normalized Hessian trace}, defined as follows:
\begin{equation}\label{eq:normHtr}
\mathcal{C}(\boldsymbol{w})=\frac{\operatorname{Tr}\left(\nabla^2 L_{\mathcal{S}}\left(\boldsymbol{w}\right)\right)}{\|\nabla L_{\mathcal{S}}(\boldsymbol{w})\|_2}
\end{equation}
This metric exhibits continual growth during SAM training, indicating increasing curvature. This behavior aligns well with the decreasing AR of one-step gradient ascent, as depicted in \fref{fig:CR}. An additional advantage of the normalized Hessian trace is its consistent trends and values across both training and test sets. In contrast, plain $\operatorname{Tr}\left(\nabla^2 L_{\mathcal{S}}\left(\boldsymbol{w}\right)\right)$ display inconsistent behaviors between these sets, as evidenced in \fref{fig:CR} (a,b,c). This discrepancy questions the viability of solely utilizing Hessian trace or the top Hessian eigenvalue for DNN regularization based on training data.

\subsection{Curvature Regularized Sharpness-Aware Minimization (CR-SAM)}
\label{subsec:crsam}
For the sake of generalization, it is preferable to steer clear of excessive non-linearity in deep learning models, as it implies highly non-convex loss surfaces. On such models, the challenge of flattening minima (which improves generalization) becomes considerably harder, potentially exceeding the capabilities of gradient-based optimizers. In this context, our proposed normalized Hessian trace \eqref{eq:normHtr} can be employed to train deep models with more manageable loss landscapes. However, a direct minimization of $\mathcal{C}(\boldsymbol{w})$ would lead to an elevation in the gradient norm $|\nabla L_{\mathcal{S}}(\boldsymbol{w})|_2$, which could adversely affect generalization \cite{zhao2022penalizing}.
Therefore, we propose to optimize $\operatorname{Tr}\left(\nabla^2 L_{\mathcal{S}}\left(\boldsymbol{w}\right)\right)$ and $\|\nabla L_{\mathcal{S}}(\boldsymbol{w})\|_2$ separately. Specifically, we penalize both $\operatorname{Tr}\left(\nabla^2 L_{\mathcal{S}}\left(\boldsymbol{w}\right)\right)$ and $\|\nabla L_{\mathcal{S}}(\boldsymbol{w})\|_2$ but with different extent such that they jointly lead to a smaller $\mathcal{C}(\boldsymbol{w})$. Thus, we introduce our proposed curvature regularizer as:
\begin{equation}
R_c(\boldsymbol{w})= \alpha \log \operatorname{Tr}\left(\nabla^2 L_{\mathcal{S}}\left(\boldsymbol{w}\right)\right) + \beta \log \|\nabla L_{\mathcal{S}}(\boldsymbol{w})\|_2
\end{equation}
where $\alpha > \beta > 0$ such that the numerator of $\mathcal{C}(\boldsymbol{w})$ is penalized more than the denominator. This regularizer is equivalent to $\alpha \log \mathcal{C}(\boldsymbol{w}) + (\alpha+\beta)\log \|\nabla L_{\mathcal{S}}(\boldsymbol{w})\|_2$, which is a combination of normalized Hessian trace with gradient norm penalty regularizer. Our regularization strategy can also be justified by analyzing the sharpness:
\begin{equation*}
\begin{split}
&R^{\mathrm{True}}(\boldsymbol{w})=\max _{\|\boldsymbol{v}\|_2 \leq 1} L_{\mathcal{S}}(\boldsymbol{w}+\rho \boldsymbol{v})-L_{\mathcal{S}}(\boldsymbol{w})\\
&=\max _{\|\boldsymbol{v}\|_2 \leq 1}\left(\rho \boldsymbol{v}^{\top} \nabla L_{\mathcal{S}}(\boldsymbol{w})+\frac{\rho^2}{2} \boldsymbol{v}^{\top} \nabla^2 L_{\mathcal{S}}(\boldsymbol{w}) \boldsymbol{v}+O\left(\rho^3\right)\right)
\end{split}
\end{equation*}

We can see that $\max _{\|\boldsymbol{v}\|_2 \leq 1} \rho \boldsymbol{v}^{\top}\nabla L_{\mathcal{S}}(\boldsymbol{w})=\rho\|\nabla L_{\mathcal{S}}(\boldsymbol{w})\|_2$ 
(cf. \eqref{eq:sam-approx}). Under the condition that $\boldsymbol{v} \sim N(0, I)$, we have $\mathbb{E}_{\boldsymbol{v} \sim N(0, I)} \boldsymbol{v}^{\top} \nabla^2 L_{\mathcal{S}}(\boldsymbol{w}) \boldsymbol{v} = \operatorname{Tr}\left(\nabla^2 L_{\mathcal{S}}\left(\boldsymbol{w}\right)\right)$ for the second term. However, the first-order term $\|\nabla L_{\mathcal{S}}(\boldsymbol{w})\|_2$ vanishes at the local minimizers of the loss $L$, and thus the second-order term will become prominent and hence be penalized. Therefore, introducing our regularizer will have the effect of penalizing both the Hessian trace and the gradient norm and thereby reduce the sharpness of a loss landscape.

Informed by our heuristic and theoretical analysis above, our CR-SAM optimizes the following objective:
\begin{align}
\min _{\boldsymbol{w}} &L^{\operatorname{CR-SAM}}(\boldsymbol{w}) \nonumber \\
\text { where } & L^{\operatorname{CR-SAM}}(\boldsymbol{w}) = L^{\operatorname{SAM}}(\boldsymbol{w}) + R_c(\boldsymbol{w})
\end{align}

\subsection{Solving Computational Efficiency}
\label{subsec:fd}

Computing the Hessian trace as in $R_c(\boldsymbol{w})$ for very large matrices is computationally intensive, especially for modern over-parameterized DNNs with millions of parameters. To address this issue, we first propose a stochastic estimators for $R_c(\boldsymbol{w})$:
\begin{equation*}
\begin{split}
    &R_c(\boldsymbol{w}) = \alpha \log \operatorname{Tr}\left(\nabla^2 L_{\mathcal{S}}\left(\boldsymbol{w}\right)\right) + \beta \log \|\nabla L_{\mathcal{S}}(\boldsymbol{w})\|_2 \\
    &= \mathbb{E}_{\boldsymbol{v} \sim N(0, I)} \left[\alpha \log \boldsymbol{v}^{\top} \nabla^2 L_{\mathcal{S}}(\boldsymbol{w}) \boldsymbol{v}+ \beta \log \boldsymbol{v}^{\top} \nabla L_{\mathcal{S}}(\boldsymbol{w})\right]
\end{split}
\end{equation*}
which reduces Hessian trace computation to averages of Hessian-vector products. However, the complexity of computing the Hessian-vector products in the above estimator is still high for optimizers in large scale problems. Hence, we further propose an approximation based on finite difference (FD) which not only reduces the computational complexity, but also makes the computation \emph{parallelizable}.

\begin{theorem} \label{theorem:fd}
If $L_{\mathcal{S}}(\boldsymbol{w})$ is 2-times-differentiable at $\boldsymbol{w}$, with $\boldsymbol{v} \sim N(0, I)$ , by finite difference we have
\begin{equation*}
\left\{\begin{array}{l}
\hspace{-0.2cm}\boldsymbol{v}^{\top} \nabla L_{\mathcal{S}}(\boldsymbol{w}) =\frac{1}{2\rho} (L_{\mathcal{S}}(\boldsymbol{w} + \rho\boldsymbol{v})- L_{\mathcal{S}}(\boldsymbol{w} - \rho\boldsymbol{v}))+o\left(\epsilon^2\right) ; \\

\begin{aligned}
\hspace{-0.2cm}\boldsymbol{v}^{\top} \nabla^2 L_{\mathcal{S}}(\boldsymbol{w}) \boldsymbol{v} &= \frac{1}{\rho^2} (L_{\mathcal{S}}(\boldsymbol{w}+\rho\boldsymbol{v})+L_{\mathcal{S}}(\boldsymbol{w}-\rho\boldsymbol{v}) \\
&-2 L_{\mathcal{S}}(\boldsymbol{w}))+o\left(\epsilon^3\right).
\end{aligned}
\end{array}\right.
\end{equation*}
\end{theorem}

By Theorem \ref{theorem:fd}, we can instantiate $R_c(\boldsymbol{w})$ as:
\begin{align}\label{eq:R_c_fd}
&R_c(\boldsymbol{w}) =\mathbb{E}_{\boldsymbol{v} \sim N(0, I)} \Big[\alpha \log\big(L_{\mathcal{S}}(\boldsymbol{w}+\rho\boldsymbol{v})+L_{\mathcal{S}}(\boldsymbol{w}-\rho\boldsymbol{v}) \notag\\ 
&-2 L_{\mathcal{S}}(\boldsymbol{w})\big) 
+ \beta \log\big( L_{\mathcal{S}}(\boldsymbol{w} + \rho\boldsymbol{v})-L_{\mathcal{S}}(\boldsymbol{w} - \rho\boldsymbol{v})\big)\Big] \notag\\
&+ const.
\end{align}

\begin{figure*}
\begin{minipage}{.45\textwidth}
\includegraphics[width=.9\textwidth]{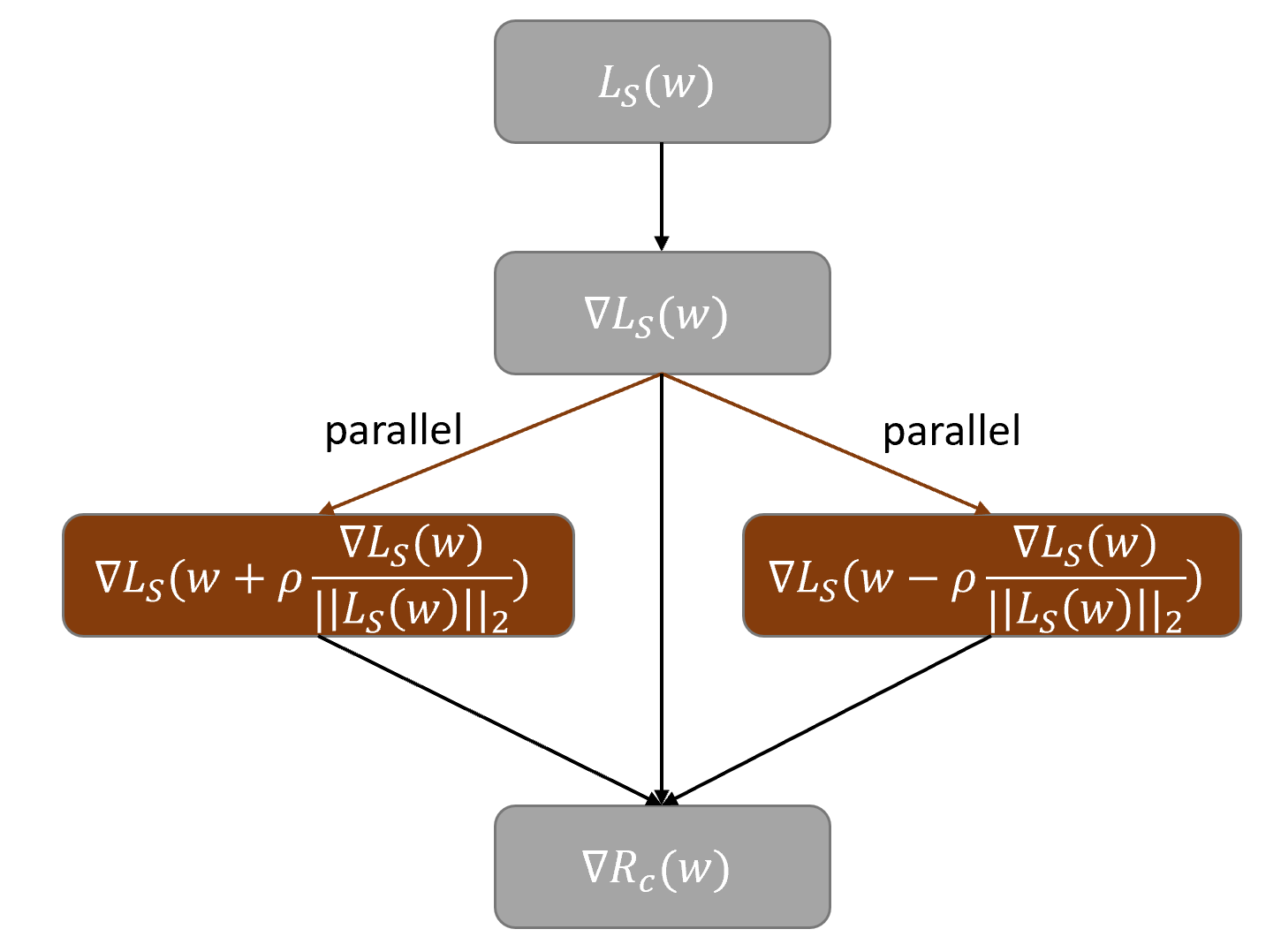}
\caption{Computing the gradient of $R_c(\boldsymbol{w})$. The two gradient steps are independent of each other and can be perfectly parallelized. Hence the training speed is almost the same as SAM.}
\label{fig:computation}
\end{minipage}
\hfill
\begin{minipage}{0.5\textwidth}
\begin{algorithm}[H]
\caption{Training with CR-SAM}
\label{alg:CR-SAM}
\begin{flushleft}
    \textbf{Input:} Training set $\mathcal{S}$; DNN model $f(x ; \boldsymbol{w})$; Loss funtion $\ell\left(f\left(x_i ; \boldsymbol{w}\right), y_i\right)$; Batch size $B$; Learning rate $\eta$; Perturbation size $\rho$; regularizer coefficients $\alpha$ and $\beta$ \\
    \textbf{Output:} model trained by CR-SAM
\end{flushleft}
\begin{algorithmic}[1]
    \State Parameter initialization $\boldsymbol{w}_0$.
    \While{\emph{not converged}}
        \State Sample batch $\mathcal{B}=\left\{\left({x}_i, {y}_i\right)\right\}_{i=0}^B$ from $\mathcal{S}$;
        \State Compute $\boldsymbol{v} = \frac{\nabla L_{\mathcal{S}}(\boldsymbol{w})}{\|\nabla L_{\mathcal{S}}(\boldsymbol{w})\|_2}$;
        \State Compute $L_{\mathcal{S}}(\boldsymbol{w}+\rho\boldsymbol{v})$ and $L_{\mathcal{S}}(\boldsymbol{w}-\rho\boldsymbol{v})$ ;
        \State Compute $\nabla R_c(\boldsymbol{w})$ per equation \ref{eq:R_c_fd};
        \State $\boldsymbol{w}_{t+1} = \boldsymbol{w}_t-\eta (\nabla\mathcal{L}\left(\boldsymbol{w}_t\right)+R_c\left(\boldsymbol{w}_t\right))$;
    \EndWhile
    \State \textbf{return} $\boldsymbol{w}_t$
\end{algorithmic}
\end{algorithm}
\end{minipage}
\vspace{-0.5cm}
\end{figure*}

The above formulation involves an expectation over $\boldsymbol{v}$, which uniformly penalizes expected curvature across all directions. Previous studies \cite{fawzi2018empirical,moosavi2019robustness} highlight that gradient directions represent high-curvature directions. Hence, we choose to optimize over perturbations solely along gradient directions, approximating $R_c(\boldsymbol{w})$ by considering $\boldsymbol{v} = \nabla L_{\mathcal{S}}(\boldsymbol{w})$. Additionally, the terms $L_{\mathcal{S}}(\boldsymbol{w} + \rho\boldsymbol{v})$ and $L_{\mathcal{S}}(\boldsymbol{w} - \rho\boldsymbol{v})$ can be computed in parallel as shown in \fref{fig:computation}.

We offer a meaningful interpretation of the finite difference regularizer \eqref{eq:R_c_fd}: The second term within $R_c(\boldsymbol{w})$, i.e., $[L_{\mathcal{S}}(\boldsymbol{w} + \rho\boldsymbol{v}) - L_{\mathcal{S}}(\boldsymbol{w} - \rho\boldsymbol{v})]$, resembles the surrogate gap $[L_{\mathcal{S}}(\boldsymbol{w} + \rho\boldsymbol{v}) - L_{\mathcal{S}}(\boldsymbol{w})]$ as introduced in \cite{zhuang2022surrogate}. However, unlike solely focusing on optimizing the ridge (locally worst-case perturbation) within the $\rho$-bounded neighborhood around the current parameter vector, our proposed regularizer also delves into the valley (locally best-case perturbation) of the DNN loss landscape, with their loss discrepancies similarly constrained by $R_c(\boldsymbol{w})$. Additionally, by expressing the first term within $R_c(\boldsymbol{w})$ as $[L_{\mathcal{S}}(\boldsymbol{w}+\rho\boldsymbol{v}) - L_{\mathcal{S}}(\boldsymbol{w})] - [L_{\mathcal{S}}(\boldsymbol{w}) - L_{\mathcal{S}}(\boldsymbol{w}-\rho\boldsymbol{v})]$, our approach encourages minimizing the disparity between the worst-case perturbed sharpness and the best-case perturbed sharpness. In essence, our strategy jointly optimizes the worst-case and best-case perturbations within the parameter space neighborhood, promoting a smoother, flatter loss landscape with fewer excessive wavy ridges and valleys.


The full pseudo-code of our CR-SAM training is given in Algorithm~\ref{alg:CR-SAM}.

\section{Experiments}
\label{sec:experiments}

To assess CR-SAM, we conduct thorough experiments on prominent image classification benchmark datasets: CIFAR-10/CIFAR-100 and ImageNet-1k/-C/-R. Our evaluation encompasses a wide array of network architectures, including ResNet, WideResNet, PyramidNet, and Vision Transformer (ViT), in conjunction with diverse data augmentation techniques. These experiments are implemented using PyTorch and executed on Nvidia A100 and V100 GPUs.

\subsection{Training from Scratch on CIFAR-10 / CIFAR-100}
\label{subsec: cifar}

\textbf{Setup.} 
In this section, we evaluate CR-SAM using the CIFAR-10/100 datasets \cite{krizhevsky2009learning}. Our evaluation encompasses a diverse selection of widely-used DNN architectures with varying depths and widths. Specifically, we employ ResNet-18 \cite{he2016deep}, ResNet-50 \cite{he2016deep}, Wide ResNet-28-10 (WRN-28-10) \cite{zagoruyko2016wide}, and PyramidNet-110 \cite{han2017deep}, along with a range of data augmentation techniques, including basic augmentations (horizontal flip, padding by four pixels, and random crop) \cite{foret2021sharpnessaware}, Cutout \cite{devries2017improved}, and AutoAugment \cite{cubuk2018autoaugment}, to ensure a comprehensive assessment.

Following the setup in \cite{liu2022random, du2022sharpness}, we train all models from scratch for 200 epochs, using batch size 128 and employing a cosine learning rate schedule. We conduct grid search to determine the optimal learning rate, weight decay, perturbation magnitude ($\rho$), coefficient ($\alpha$ and $\beta$) values that yield the highest test accuracy. To ensure a fair comparison, we run each experiment three times with different random seeds. Further details of the setup are provided in Appendix.

\textbf{Results.}
Refer to Table \ref{tab:cifar} for a comprehensive overview. CR-SAM consistently outperforms both vanilla SAM and SGD across all configurations on both CIFAR-10 and CIFAR-100 datasets. Notable improvements are observed, such as a 1.11\% enhancement on CIFAR-100 with ResNet-18 employing cutout augmentation and a 1.30\% boost on CIFAR-100 with WRN-28-10 using basic augmentation. 

Furthermore, we empirically observe that CR-SAM exhibits a faster convergence rate in comparison to vanilla SAM (details in Appendix). This accelerated convergence could be attributed to CR-SAM's ability to mitigate excessive curvature, ultimately reducing optimization complexity and facilitating swifter arrival at local minima.

\begin{table*}[h]
    \centering
    \small
    \caption{Results on CIFAR-10 and CIFAR-100. The base optimizer for
SAM and CR-SAM is SGD with Momentum (SGD+M).}
    \resizebox{0.9\linewidth}{!}{
    \begin{tabular}{c|c|ccc|ccc}
        \toprule
        & & \multicolumn{3}{c|}{CIFAR-10} & \multicolumn{3}{c}{CIFAR-100} \\
        \midrule
        Model & Aug & SGD  & SAM & CR-SAM  & SGD & SAM & CR-SAM  \\
        \midrule
        & Basic & 95.29$_{\pm 0.16}$ & 96.46$_{\pm 0.18}$ & \textbf{96.95}$_{\pm 0.13}$ & 78.34$_{\pm 0.22}$  & 79.81$_{\pm 0.18}$ & \textbf{80.76}$_{\pm 0.21}$ \\
        ResNet-18 & Cutout & 95.96$_{\pm 0.13}$ & 96.55$_{\pm 0.15}$ & \textbf{97.01}$_{\pm 0.21}$ & 79.23$_{\pm 0.13}$ &  80.15$_{\pm 0.17}$ & \textbf{81.26}$_{\pm 0.19}$ \\
        & AA & 96.33$_{\pm 0.15}$ &  96.75$_{\pm 0.18}$ & \textbf{97.27}$_{\pm 0.12}$ & 79.05$_{\pm 0.17}$ & 81.26$_{\pm 0.21}$ & \textbf{82.11}$_{\pm 0.22}$ \\
        \midrule
        & Basic & 96.35$_{\pm 0.12}$ & 96.51$_{\pm 0.16}$ & \textbf{97.14}$_{\pm 0.11}$ & 80.54$_{\pm 0.13}$& 82.11$_{\pm 0.12}$ & \textbf{83.03}$_{\pm 0.17}$ \\
        ResNet-101 & Cutout & 96.56$_{\pm 0.18}$ & 96.95$_{\pm 0.13}$ & \textbf{97.51}$_{\pm 0.24}$ & 81.26$_{\pm 0.21}$  & 82.39$_{\pm 0.27}$ & \textbf{83.46}$_{\pm 0.16}$ \\
        & AA & 96.78$_{\pm 0.14}$ & 97.11$_{\pm 0.16}$ & \textbf{97.76}$_{\pm 0.16}$ & 81.83$_{\pm 0.37}$ & 83.25$_{\pm 0.47}$ & \textbf{84.19}$_{\pm 0.23}$ \\
        \midrule
        & Basic & 95.89$_{\pm 0.21}$ & 96.81$_{\pm 0.26}$ & \textbf{97.36}$_{\pm 0.15}$ & 81.84$_{\pm 0.13}$  & 83.15$_{\pm 0.14}$ & \textbf{84.45}$_{\pm 0.09}$ \\
        WRN-28-10 & Cutout & 96.89$_{\pm 0.07}$  & 97.55$_{\pm 0.16}$ & \textbf{97.98}$_{\pm 0.21}$ & 81.96$_{\pm 0.40}$  & 83.47$_{\pm 0.15}$ & \textbf{84.48}$_{\pm 0.13}$ \\
        & AA & 96.93$_{\pm 0.12}$   & 97.59$_{\pm 0.06}$ & \textbf{97.94}$_{\pm 0.08}$ & 82.16$_{\pm 0.11}$  & 83.69$_{\pm 0.26}$ & \textbf{84.74}$_{\pm 0.21}$ \\
        \midrule
        & Basic & 96.27$_{\pm 0.13}$  & 97.34$_{\pm 0.13}$ & \textbf{97.89}$_{\pm 0.08}$ & 83.27$_{\pm 0.12}$  & 84.89$_{\pm 0.09}$ & \textbf{85.68}$_{\pm 0.14}$ \\
        PyramidNet-110 & Cutout & 96.79$_{\pm 0.13}$  & 97.61$_{\pm 0.21}$ & \textbf{98.08}$_{\pm 0.11}$ & 83.43$_{\pm 0.21}$  & 84.97$_{\pm 0.17}$ & \textbf{85.86}$_{\pm 0.21}$ \\
        & AA & 96.97$_{\pm 0.08}$  & 97.81$_{\pm 0.13}$ & \textbf{98.26}$_{\pm 0.11}$ & 84.59$_{\pm 0.08}$  & 85.76$_{\pm 0.23}$ & \textbf{86.58}$_{\pm 0.14}$ \\
        \bottomrule
    \end{tabular}}
    \label{tab:cifar}
    \vspace{-0.3cm}
\end{table*}

\subsection{Training from Scratch on ImageNet-1k/-C/-R}
\label{subsec:imagenet}

\textbf{Setup.} 
This section details our evaluation on the ImageNet dataset \cite{deng2009imagenet}, containing 1.28 million images across 1000 classes. We assess the performance of ResNet \cite{he2016deep} and Vision Transformer (ViT) \cite{dosovitskiy2020image} architectures. Evaluation is extended to out-of-distribution data, namely ImageNet-C \cite{hendrycks2018benchmarking} and ImageNet-R \cite{hendrycks2021many}. ResNet50, ResNet101, ViT-S/32, and ViT-B/32 are evaluated with Inception-style preprocessing.

For ResNet models, SGD serves as the base optimizer. We follow the setup in \cite{du2022efficient}, training ResNet50 and ResNet101 with batch size 512 for 90 epochs. The initial learning rate is set to 0.1, progressively decayed using a cosine schedule. For ViT models, we adopt AdamW \cite{loshchilov2018decoupled} as the base optimizer with parameters $\beta_1=0.9$ and $\beta_2=0.999$. ViTs are trained with batch size 512 for 300 epochs. Refer to the Appendix for further training specifics.

\textbf{Results.}
Summarized in Table \ref{tab:imagenet}, our results indicate substantial accuracy improvements across various DNN models, including ResNet and ViT, on the ImageNet dataset. Notably, CR-SAM's performance surpasses that of SAM by 1.16\% for ResNet-50 and by 1.77\% for ViT-B/32. These findings underscore the efficacy of our CR-SAM approach.

\begin{table}[h]
    \centering
    \small
    \caption{Results on ImageNet-1k/-C/-R, the base optimizer for ResNets and ViTs
are SGD+M and AdamW, respectively.}
    \resizebox{0.98\linewidth}{!}{
    \begin{tabular}{c|c|cccc}
        \toprule
         Model & Datasets & Vanilla & SAM & R-SAM & CR-SAM  \\
        \midrule
        & ImageNet-1k & 75.94 & 76.48 & 76.89 &\textbf{77.64}  \\
        ResNet-50 & ImageNet-C & 43.64 & 46.03 & 46.19 &\textbf{46.94}  \\
        & ImageNet-R & 21.93 & 23.13 & 22.89 &\textbf{23.48}  \\
        \midrule
        & ImageNet-1k & 77.81 & 78.64 & 78.71 &\textbf{79.12} \\
        ResNet-101 & ImageNet-C & 48.56  & 51.27 & 51.35 &\textbf{51.87}  \\
        & ImageNet-R & 24.38 & 25.89 & 25.91 &\textbf{26.37}  \\
        \midrule
        \midrule
        & ImageNet-1k & 68.40 & 70.23 & 70.39 &\textbf{71.68} \\
        ViT-S/32 & ImageNet-C & 43.21 & 45.78 & 45.92 &\textbf{46.46}  \\
        & ImageNet-R & 19.04 & 21.12 & 21.35 &\textbf{21.98}  \\
        \midrule
        & ImageNet-1k & 71.25 & 73.51 & 74.06 &\textbf{75.28} \\
        ViT-B/32 & ImageNet-C & 44.37  & 46.98 & 47.28&\textbf{48.12}  \\
        & ImageNet-R & 23.12 & 24.31 & 24.53 &\textbf{25.04}  \\
        \bottomrule
    \end{tabular}}
    \vspace{-0.1cm}
    \label{tab:imagenet}
\end{table}

\subsection{Model Geometry Analysis}
\label{subsec:eigenvalue}
CR-SAM aims to reduce the normalized trace of the Hessian to promote flatter minima. Empirical validation of CR-SAM's ability to locate optima with lower curvature is presented through model geometry comparisons among models trained by SGD, SAM, and CR-SAM (see Table \ref{tab:geometry}). Our analysis is based on ResNet-18 trained on CIFAR-100 for 200 epochs using the three optimization methods. Hutchinson’s method \cite{avron2011randomized, yao2020pyhessian} is utilized to compute the Hessian trace, with values obtained from the test set across three independent runs. Notably, the results reveal that CR-SAM significantly reduces both gradient norms and Hessian traces throughout training in contrast to SGD and SAM. This reduction contributes to a smaller normalized Hessian trace, affirming the effectiveness of our proposed regularization strategy.


\begin{table}[h]
  \begin{center}
    \caption{Model geometry of ResNet-18 models trained with SGD, SAM and CR-SAM, values are computed on test set.}
    \label{tab:geometry}
    \resizebox{0.98\linewidth}{!}{
    \begin{tabular}{c | c | c | c | c} 
     \midrule
     \textbf{Optimizer} & $\|\nabla L_{\mathcal{S}}(\boldsymbol{w})\|_2$ & $\operatorname{Tr}\left(\nabla^2 L_{\mathcal{S}}\left(\boldsymbol{w}\right)\right)$ & $\mathcal{C}(\boldsymbol{w})$ & Accuracy (\%) \\ 
     \midrule
     SGD & 19.97 $_{\pm 0.52}$ & 32673.88 $_{\pm 1497.56}$ & 1674.89 $_{\pm 78.69}$ & 78.34 $_{\pm 0.22}$   \\ 
     SAM & 11.51 $_{\pm 0.31}$ & 14176.52 $_{\pm 327.69}$ & 1193.87 $_{\pm 59.18}$ & 79.81 $_{\pm 0.18}$\\
     CR-SAM & \textbf{8.26 $_{\pm 0.19}$ }& \textbf{7968.19 $_{\pm 145.73}$} & \textbf{884.95 $_{\pm 23.59}$} & \textbf{80.76 $_{\pm 0.21}$}  \\ 
     \midrule
    \end{tabular}}
    \vspace{-0.5cm}
  \end{center}
  \end{table}

\subsection{Visualization of Landscapes}
\label{subsec:viz}
\begin{figure}[h]

    \centering
    \begin{subfigure}[b]{0.27\linewidth}
        \centering
        \includegraphics[width=0.8\linewidth]{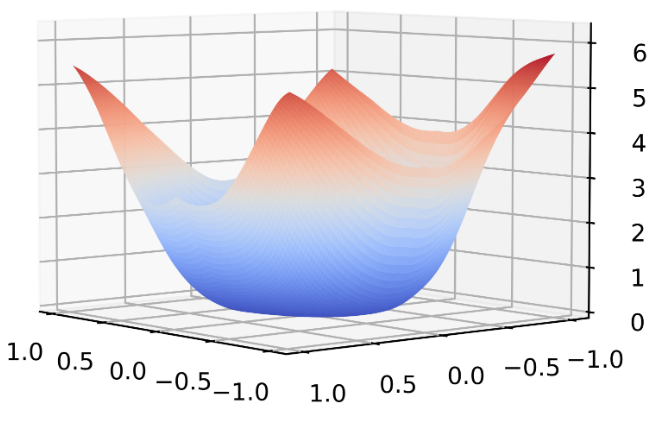}
        \vspace{-5px}
        \caption{SGD}
        \label{fig:sgd}
    \end{subfigure}
    \hfill
    \begin{subfigure}[b]{0.3\linewidth}
        \centering
        \includegraphics[width=0.8\linewidth]{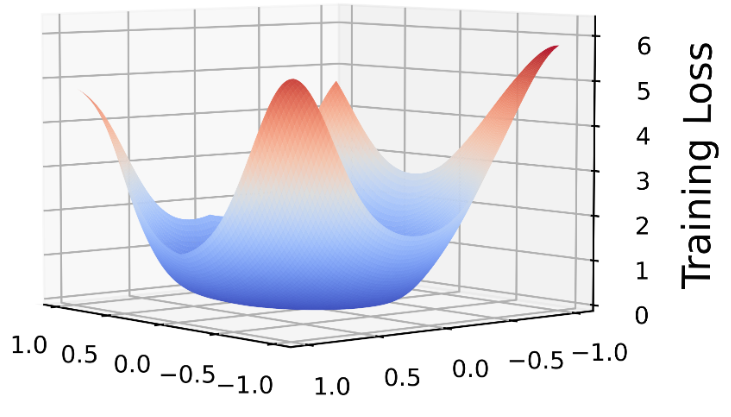}
        \vspace{-5px}
        \caption{SAM}
        \label{fig:sam}
    \end{subfigure}
    \hfill
    \begin{subfigure}[b]{0.3\linewidth}
        \centering
        \includegraphics[width=0.8\linewidth]{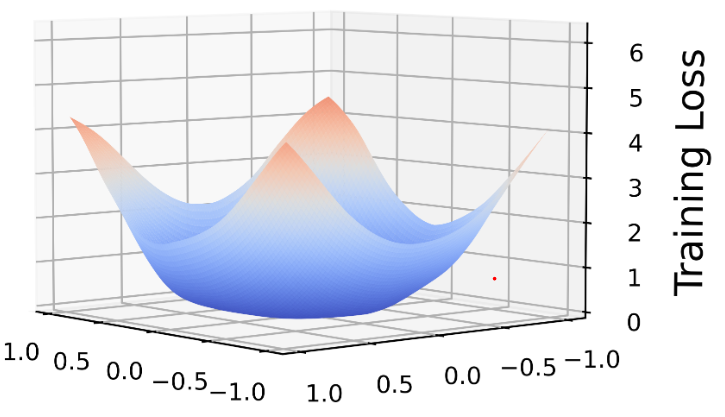}
        \vspace{-5px}
        \caption{CR-SAM}
        \label{fig:sam-GAM}
    \end{subfigure}
    \caption{Loss landscapes for SGD, SAM and CR-SAM.}
    \label{fig:vis}
    \vspace{-0.5cm}
\end{figure}

We visualize the flatness of minima obtained using CR-SAM by plotting loss landscapes of PyramidNet110 trained with SGD, SAM, and CR-SAM on CIFAR-100 for 200 epochs. Employing the visualization techniques from \cite{li2018visualizing}, we depict loss values along two randomly sampled orthogonal Gaussian perturbations around local minima. As depicted in \fref{fig:vis}, the visualization illustrates that CR-SAM yields flatter minima compared to SGD and SAM.

\section{Conclusion}
\label{sec:conclusion}
In this paper, we identify the limitations of the one-step gradient ascent in SAM's inner maximization during training due to the excessive non-linearity of the loss landscape. In addition, existing curvature metrics lack the ability to precisely capture the loss function geometry. To address these issues, we introduce normalized Hessian trace, which offers consistent and accurate characterization of loss function curvature on both training and test data. Building upon this tool, we present CR-SAM, a novel training approach for enhancing neural network generalizability by regularizing our proposed curvature metric. Additionally, to mitigate the overhead of computing the Hessian trace, we incorporate a parallelizable finite difference method. Our comprehensive experiments that span a wide variety of model architectures across popular image classification datasets including CIFAR10/100 and ImageNet-1k/-C/-R, affirm the effectiveness of our proposed CR-SAM training strategy. We hypothesize that combining our proposed regularizer with other SAM variants would be a promising direction toward enhanced DNN models.

\section{Acknowledgement}
This work was supported in part by the National Science Foundation (NSF) under Grant No. 2008878, and in part by the Air Force Research Laboratory (AFRL) and the Lifelong Learning Machines program by DARPA/MTO under Contract No. FA8650-18-C-7831. 
The research was also sponsored by the Army Research Laboratory and was accomplished under Cooperative Agreement Number W911NF-22-2-0209. 

\bibliography{aaai24}

\newpage

\section{Proofs}
In this section we provide proofs for Theorem 1 and Theorem 2 in the main text.

\subsection{Proof of Theorem 1.}
\begin{theorem} \label{theorem:fd}
If $L_{\mathcal{S}}(\boldsymbol{w})$ is 2-times-differentiable at $\boldsymbol{w}$, with $\boldsymbol{v} \sim N(0, I)$ , by finite difference we have
\begin{equation*}
\left\{\begin{array}{l}
\hspace{-0.2cm}\boldsymbol{v}^{\top} \nabla L_{\mathcal{S}}(\boldsymbol{w}) =\frac{1}{2\rho} (L_{\mathcal{S}}(\boldsymbol{w} + \rho\boldsymbol{v})- L_{\mathcal{S}}(\boldsymbol{w} - \rho\boldsymbol{v}))+o\left(\epsilon^2\right) ; \\

\begin{aligned}
\hspace{-0.2cm}\boldsymbol{v}^{\top} \nabla^2 L_{\mathcal{S}}(\boldsymbol{w}) \boldsymbol{v} &= \frac{1}{\rho^2} (L_{\mathcal{S}}(\boldsymbol{w}+\rho\boldsymbol{v})+L_{\mathcal{S}}(\boldsymbol{w}-\rho\boldsymbol{v}) \\
&-2 L_{\mathcal{S}}(\boldsymbol{w}))+o\left(\epsilon^3\right).
\end{aligned}
\end{array}\right.
\end{equation*}
\end{theorem}

\textbf{Proof}: Using Taylor polynomial expansion of $L_{\mathcal{S}}(\boldsymbol{w}+\rho\boldsymbol{v})$ and $L_{\mathcal{S}}(\boldsymbol{w}-\rho\boldsymbol{v})$ centered at $\boldsymbol{w}$. We have

\begin{equation}
\left\{\begin{array}{l}
L_{\mathcal{S}}(\boldsymbol{w}+\rho\boldsymbol{v})=L_{\mathcal{S}}(\boldsymbol{w}) + \rho\boldsymbol{v}\nabla L_{\mathcal{S}}(\boldsymbol{w})+\mathcal{O}\left(\rho^2\right) ; \\[10pt]
L_{\mathcal{S}}(\boldsymbol{w}-\rho\boldsymbol{v})=L_{\mathcal{S}}(\boldsymbol{w}) - \rho\boldsymbol{v}\nabla L_{\mathcal{S}}(\boldsymbol{w})+\mathcal{O}\left(\rho^2\right) .
\end{array}\right.
\end{equation}

Thus rearranging the above two qwuation we can obtain $\boldsymbol{v}^{\top} \nabla L_{\mathcal{S}}(\boldsymbol{w}) =\frac{1}{2\rho} (L_{\mathcal{S}}(\boldsymbol{w} + \rho\boldsymbol{v})- L_{\mathcal{S}}(\boldsymbol{w} - \rho\boldsymbol{v}))+O\left(\rho^2\right)$.

We rewrite $\boldsymbol{v}^{\top} \nabla^2 L_{\mathcal{S}}(\boldsymbol{w}) \boldsymbol{v} $ as directional derivatives as $\nabla_{\boldsymbol{v}}^{2} L_{\mathcal{S}}(\boldsymbol{w})$. Reapply the above formulation gives

\begin{align*}
    \nabla_{\boldsymbol{v}}^{2} L_{\mathcal{S}}(\boldsymbol{w}) &= \frac{1}{\rho} (\nabla_{\boldsymbol{v}} L_{\mathcal{S}}(\boldsymbol{w} + 0.5\rho\boldsymbol{v})- \nabla_{\boldsymbol{v}} L_{\mathcal{S}}(\boldsymbol{w} - 0.5\rho\boldsymbol{v})) \\
    &+O\left(\rho^2\right) \\
    &= \frac{1}{\rho^2} (L_{\mathcal{S}}(\boldsymbol{w} + 0.5\rho\boldsymbol{v}+ 0.5\rho\boldsymbol{v}) - L_{\mathcal{S}}(\boldsymbol{w}\\
    & + 0.5\rho\boldsymbol{v}- 0.5\rho\boldsymbol{v}) - L_{\mathcal{S}}(\boldsymbol{w} - 0.5\rho\boldsymbol{v} + 0.5\rho\boldsymbol{v}) \\
    &+ L_{\mathcal{S}}(\boldsymbol{w} - 0.5\rho\boldsymbol{v} - 0.5\rho\boldsymbol{v}))+O\left(\rho^2\right) \\
    &= \frac{1}{\rho^2} (L_{\mathcal{S}}(\boldsymbol{w} + \rho\boldsymbol{v})+ L_{\mathcal{S}}(\boldsymbol{w} -\rho\boldsymbol{v})- 2L_{\mathcal{S}}(\boldsymbol{w})) \\
    &+O\left(\rho^2\right)
\end{align*}

\qed

\subsection{Proof of PAC-Bayesian generalization error bounds.}
\begin{theorem} \label{theorem:bound}
(Stated informally) For any $\delta \in(0,1)$, with probability at least $1-\delta$ over a draw of the training set $\mathcal{S}$ and a solution $\boldsymbol{w^*}$ found by a gradient-based optimizer, by picking $\boldsymbol{v}$ follows standard Gaussian, the following inequality holds:
\begin{align*} 
\mathbb{E}_{\boldsymbol{v} \sim N(0, I)} L_{\mathcal{D}}(\boldsymbol{w^*}+\rho \boldsymbol{v}) &\leq L_{\mathcal{S}}(\boldsymbol{w^*}) + \frac{\rho^2}{2} \operatorname{Tr}\left(\nabla^2 L\left(\boldsymbol{w^*}\right)\right) \\
& + \gamma\|\nabla L_{\mathcal{S}}(\boldsymbol{w^*})\|_2  + 
h\left(\|\boldsymbol{w^*}\|_2^2 / \rho^2\right) \nonumber
\end{align*}
\end{theorem}

\textbf{Proof sketch}: 
We follow the most basic PAC-Bayesian generalization error bounds as \cite{JMLR:v17:15-290, gat2022importance}: For any $\lambda>0$, for any $\delta \in(0,1)$ and for any prior distribution $p$, with probability at least $1-\delta$ over the draw of the training set $S$, the following holds simultaneously for any posterior distribution $q$ :

\begin{equation}
\mathbb{E}_{w \sim q}\left[L_D(w)\right] \leq \mathbb{E}_{w \sim q}\left[L_\mathcal{S}(w)\right]+\frac{1}{\lambda}[C(\lambda, p)+K L(q \| p)+\log (1 / \delta)]
\end{equation}

where $C(\lambda, p) \triangleq \log \left(\mathbb{E}_{w \sim p}\left[e^{\lambda\left(L_D(w)-L_S(w)\right)}\right]\right)$.

We can rearrange the first term as 
\begin{align*}
\mathbb{E}_{w \sim q}\left[L_\mathcal{S}(w)\right] &= L_{\mathcal{S}}(\boldsymbol{w}) 
 + \mathbb{E}_{w \sim q}\left[L_\mathcal{S}(w)\right] - L_{\mathcal{S}}(\boldsymbol{w}) \\
 &= L_{\mathcal{S}}(\boldsymbol{w}) + \underset{\boldsymbol{\boldsymbol{v}} \sim \mathcal{N}(\mathbf{0}, I)}{\mathbb{E}}\left[L_{\mathcal{S}}(\boldsymbol{w}+\rho\boldsymbol{v})\right]-L_{\mathcal{S}}(\boldsymbol{w}) \\
 &= L_{\mathcal{S}}(\boldsymbol{w}) + \frac{\rho^2}{2} \operatorname{Tr}\left(\nabla^2 L\left(\boldsymbol{w}\right)\right)
\end{align*}

From \cite{gat2022importance}, we have $C(\lambda, p) \leq \gamma\|\nabla L_{\mathcal{S}}(\boldsymbol{w})\|_2$ and $K L(q \| p)$ is a function of $\|\boldsymbol{w}\|_2^2$  when $\boldsymbol{v} \sim N(0, I)$

\qed

\section{Convergence Rate}

\begin{figure*}
   \centering
   \begin{subfigure}[b]{0.48\textwidth}
     \centering
     \includegraphics[width=\textwidth]{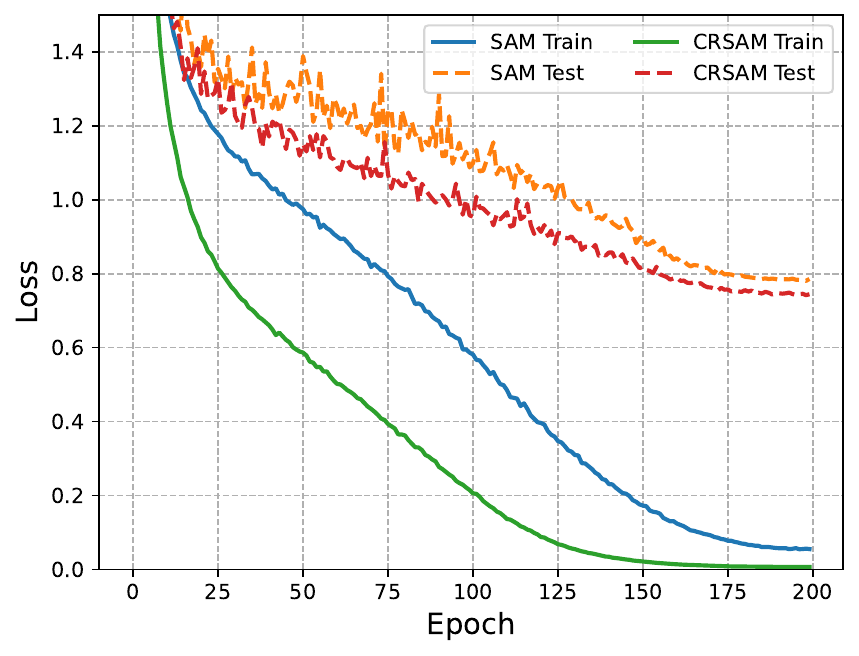}
     \caption{Loss vs Epochs of CR-SAM.}
     \label{fig:loss}
   \end{subfigure}
   \hfill
   \begin{subfigure}[b]{0.48\textwidth}
     \centering
     \includegraphics[width=\textwidth]{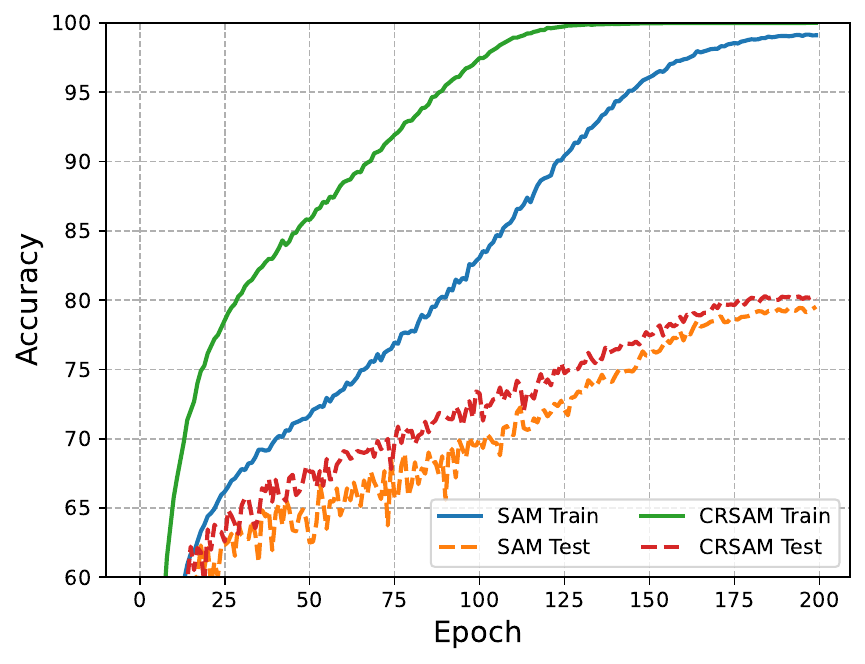}
     \caption{Accuracy vs epochs of CR-SAM.}
     \label{fig:acc}
   \end{subfigure}

    \caption{\footnotesize The evolution of training and testing loss/accuracy on CIFAR100 trained with ResNet18 by SAM and our proposed CR-SAM. The faster convergence rate of CR-SAM could be explained by the fact that CR-SAM discourages excessive curvature and thus reduces the optimization complexity, thereby making local minimum easier to reach.}
    \label{fig:convergence}
\end{figure*}

The convergence of loss in a single run of CR-SAM is presented in Figure. \ref{fig:convergence}, it shows that CR-SAM converge at the much faster rate than SAM, which indicate lesser training epocha and time.

\section{Details of Experimental Settings}

The details of experimental settings in our paper for training CIFAR10/100 and ImageNet from scratch are shown in \tref{tab:hyper_cifar} and \tref{tab:hyper_imgnet}, respectively.

\begin{table*}[h!]
\caption{\footnotesize Hyperparameters for training from scratch on CIFAR10 and CIFAR100}
  \centering
  \small
  \begin{tabular}{c|ccc|ccc}
  \toprule
     & \multicolumn{3}{c|}{\textbf{CIFAR-10 }} & \multicolumn{3}{c}{ \textbf{CIFAR-100}} \\
   \textbf{ResNet-18 }  & SGD & SAM &CR-SAM& SGD & SAM &CR-SAM\\
   \midrule
   Epoch&\multicolumn{3}{c|}{200}&\multicolumn{3}{c}{200}\\
   Batch size&\multicolumn{3}{c|}{128}&\multicolumn{3}{c}{128}\\
   Data augmentation &\multicolumn{3}{c|}{Basic}&\multicolumn{3}{c}{Basic}\\
   Peak learning rate &\multicolumn{3}{c|}{0.05}&\multicolumn{3}{c}{0.05}\\
   Learning rate decay &\multicolumn{3}{c|}{Cosine}&\multicolumn{3}{c}{Cosine}\\
   Weight decay &\multicolumn{3}{c|}{$5\times10^{-3}$}&\multicolumn{3}{c}{$5\times10^{-3}$}\\
   $\rho$ & -&0.05&0.10&-&0.10&0.15\\
   $\alpha$ &-&-&0.1&-&-&0.5\\
   $\beta$ &-&-&0.01&-&-&0.01\\
    \midrule
         \textbf{ResNet-101 }  & SGD & SAM &CR-SAM& SGD & SAM &CR-SAM\\
   \midrule
   Epoch&\multicolumn{3}{c|}{200}&\multicolumn{3}{c}{200}\\
   Batch size&\multicolumn{3}{c|}{128}&\multicolumn{3}{c}{128}\\
   Data augmentation &\multicolumn{3}{c|}{Basic}&\multicolumn{3}{c}{Basic}\\
   Peak learning rate &\multicolumn{3}{c|}{0.05}&\multicolumn{3}{c}{0.05}\\
   Learning rate decay &\multicolumn{3}{c|}{Cosine}&\multicolumn{3}{c}{Cosine}\\
   Weight decay &\multicolumn{3}{c|}{$5\times10^{-3}$}&\multicolumn{3}{c}{$5\times10^{-3}$}\\
   $\rho$ & -&0.05&0.10&-&0.10&0.15\\
   $\alpha$ &-&-&0.2&-&-&0.5\\
   $\beta$ &-&-&0.05&-&-&0.05\\
      \midrule
      \textbf{Wide-28-10}   & SGD & SAM &CR-SAM& SGD & SAM &CR-SAM\\
         \midrule
   Epoch&\multicolumn{3}{c|}{200}&\multicolumn{3}{c}{200}\\
   Batch size&\multicolumn{3}{c|}{128}&\multicolumn{3}{c}{128}\\
   Data augmentation &\multicolumn{3}{c|}{Basic}&\multicolumn{3}{c}{Basic}\\
   Peak learning rate &\multicolumn{3}{c|}{0.05}&\multicolumn{3}{c}{0.05}\\
   Learning rate decay &\multicolumn{3}{c|}{Cosine}&\multicolumn{3}{c}{Cosine}\\
   Weight decay &\multicolumn{3}{c|}{$1\times10^{-3}$}&\multicolumn{3}{c}{$1\times10^{-3}$}\\
   $\rho$ & -&0.10&0.10&-&0.10&0.15\\
   $\alpha$ &-&-&0.5&-&-&0.5\\
   $\beta$ &-&-&0.1&-&-&0.1\\
\midrule
      \textbf{PyramidNet-110}   & SGD & SAM &CR-SAM& SGD & SAM &CR-SAM\\
   \midrule
   Epoch&\multicolumn{3}{c|}{200}&\multicolumn{3}{c}{200}\\
   Batch size&\multicolumn{3}{c|}{128}&\multicolumn{3}{c}{128}\\
   Data augmentation &\multicolumn{3}{c|}{Basic}&\multicolumn{3}{c}{Basic}\\
   Peak learning rate &\multicolumn{3}{c|}{0.05}&\multicolumn{3}{c}{0.05}\\
   Learning rate decay &\multicolumn{3}{c|}{Cosine}&\multicolumn{3}{c}{Cosine}\\
   Weight decay &\multicolumn{3}{c|}{$5\times10^{-3}$}&\multicolumn{3}{c}{$5\times10^{-3}$}\\

   $\rho$ & -&0.15&0.20&-&0.15&0.20\\
   $\alpha$ &-&-&0.5&-&-&0.5\\
   $\beta$ &-&-&0.1&-&-&0.1\\
    \bottomrule
      
  \end{tabular}
  
  \label{tab:hyper_cifar}
  \vspace{-0.5em}
\end{table*}

 \begin{table*}[h!]
\caption{\footnotesize Hyperparameters for training from scratch on ImageNet}
  \centering
  \small
  \begin{tabular}{c|ccc|ccc}
  \toprule
     & \multicolumn{3}{c|}{\textbf{ResNet-50 }} & \multicolumn{3}{c}{ \textbf{ResNet-101}} \\
   \textbf{ImageNet }  & SGD & SAM &CR-SAM& SGD & SAM &CR-SAM\\
   \midrule
   Epoch&\multicolumn{3}{c|}{90}&\multicolumn{3}{c}{90}\\
   Batch size&\multicolumn{3}{c|}{512}&\multicolumn{3}{c}{512}\\
   Data augmentation &\multicolumn{3}{c|}{Inception-style}&\multicolumn{3}{c}{Inception-style}\\
   Peak learning rate &\multicolumn{3}{c|}{1.3}&\multicolumn{3}{c}{1.3}\\
   Learning rate decay &\multicolumn{3}{c|}{Cosine}&\multicolumn{3}{c}{Cosine}\\
   Weight decay &\multicolumn{3}{c}{$3\times10^{-5}$}&\multicolumn{3}{c}{$3\times10^{-5}$}\\
   $\rho$ & -&0.10&0.15&-&0.10&0.15\\
   $\alpha$ &-&-&0.1&-&-&0.2\\
   $\beta$ &-&-&0.01&-&-&0.01\\
\midrule
   & \multicolumn{3}{c|}{\textbf{ViT-S/32}} & \multicolumn{3}{c}{ \textbf{ViT-B/32}} \\
   \textbf{ImageNet }  & SGD & SAM &CR-SAM& SGD & SAM &CR-SAM\\
   \midrule
   Epoch&\multicolumn{3}{c|}{300}&\multicolumn{3}{c}{300}\\
   Batch size&\multicolumn{3}{c|}{512}&\multicolumn{3}{c}{512}\\
   Data augmentation &\multicolumn{3}{c|}{Inception-style}&\multicolumn{3}{c}{Inception-style}\\
   Peak learning rate &\multicolumn{3}{c}{$3\times10^{-3}$}&\multicolumn{3}{c}{$3\times10^{-3}$}\\
   Learning rate decay &\multicolumn{3}{c|}{Cosine}&\multicolumn{3}{c}{Cosine}\\
   Weight decay &\multicolumn{3}{c}{0.3}&\multicolumn{3}{c}{0.3}\\
   $\rho$ & -&0.05&0.10&-&0.05&0.10\\
   $\alpha$ &-&-&0.05&-&-&0.05\\
   $\beta$ &-&-&0.01&-&-&0.01\\

    \bottomrule
      
  \end{tabular}
  
  \label{tab:hyper_imgnet}
  \vspace{-0.5em}
\end{table*}

\end{document}